\documentclass[nohyperref]{article}

\usepackage{xcolor}
\usepackage{microtype}
\usepackage{graphicx}
\usepackage{subfigure}
\usepackage{booktabs} % for professional tables
\usepackage{tcolorbox}
\usepackage{listings}
\usepackage{xspace}
\usepackage{hyperref}

\usepackage[accepted]{icml2022}

\usepackage{amsmath}
\usepackage{amssymb}
\usepackage{mathtools}
\usepackage{amsthm}
\usepackage{listings}
\usepackage{comment}

\usepackage[capitalize,noabbrev]{cleveref}

\usepackage{tikz}
\usetikzlibrary{bayesnet}

\newcommand{\cascade}{cascade\xspace}
\newcommand{\cascades}{cascades\xspace}
\newcommand{\Cascades}{Cascades\xspace}

\theoremstyle{plain}

\theoremstyle{definition}

\theoremstyle{remark}

\usepackage[textsize=tiny]{todonotes}

\DeclareMathOperator*{\argmax}{\arg\!\max}

\icmltitlerunning{Language Model Cascades}

\begin{document}

\twocolumn[
% \icmltitle{Language Models are Probabilistic Inference Programs}
\icmltitle{Language Model Cascades}
% \icmltitle{Probabilistic Inference over Language Model Programs}



\icmlsetsymbol{equal}{*}

\begin{icmlauthorlist}
\icmlauthor{David Dohan}{comp}
\icmlauthor{Winnie Xu}{comp}
\icmlauthor{Aitor Lewkowycz}{x}
\icmlauthor{Jacob Austin}{comp}
\icmlauthor{David Bieber}{comp}
\icmlauthor{Raphael Gontijo Lopes}{comp}
\icmlauthor{Yuhuai Wu}{comp}
\icmlauthor{Henryk Michalewski}{comp}
\icmlauthor{Rif A. Saurous}{comp}
\icmlauthor{Jascha Sohl-dickstein}{comp}
\icmlauthor{Kevin Murphy}{comp}
\icmlauthor{Charles Sutton}{comp}
\end{icmlauthorlist}

\icmlaffiliation{comp}{Google Research, Mountain View, United States}
\icmlaffiliation{x}{Alphabet, X, the Moonshot Factory}
% \icmlaffiliation{research}{Google Research, Mountain View, United States}

\icmlcorrespondingauthor{David Dohan}{david@ddohan.com}
\icmlcorrespondingauthor{Winnie Xu}{winniexu@cs.toronto.edu}

\icmlkeywords{Machine Learning, ICML}

\vskip 0.3in
]

\newcommand{\ddohan}[1]{\textcolor{green}{{[david: #1]}}}
\newcommand{\xwinxu}[1]{\textcolor{purple}{{[winnie: #1]}}}
\newcommand{\kevin}[1]{\textcolor{red}{{[kevin: #1]}}}
\newcommand{\rif}[1]{\textcolor{brown}{{[rif: #1]}}}
\newcommand{\aitor}[1]{\textcolor{cyan}{[aitor: #1]}}
\newcommand{\henryk}[1]{\textcolor{orange}{[henryk: #1]}}
\newcommand{\charles}[1]{\textcolor{purple}{[charles: #1]}}

\urlstyle{same}



\printAffiliationsAndNotice{}  % leave blank if no need to mention equal contribution

% TODOs:
% \begin{itemize}
% % \item Tikz imag of the QTA case, with plate notation to demonst
% % \item Verifiers illustration
% % \item Settle on graphical notation (ladder? compressed ladder?)
% %\item Tool use illustration
% %\item Discuss masked modeling
% %\item Expand appendix
% \end{itemize}

\begin{abstract}
Prompted models have demonstrated impressive few-shot learning abilities.
Repeated interactions at test-time with a single model, or the composition of multiple models together, further expands capabilities. These compositions are probabilistic models, and may be expressed in the language of graphical models with random variables whose values are complex data types such as strings. Cases with control flow and dynamic structure require techniques from probabilistic programming, 
which allow implementing disparate model structures and inference strategies in a unified language.
We formalize several existing techniques from this perspective, including scratchpads / chain of thought, verifiers, STaR, selection-inference, and tool use. We refer to the resulting programs as \emph{language model \cascades}.
%Furthermore, the few shot learning abilities of large language models may be used to amortize inference across tasks.
\end{abstract}

\section{Introduction}
\label{introduction}

Language models (LMs) have demonstrated impressive few-shot learning abilities \citep{gpt3,palm}. This has led
to a number of proposals to use LMs as the basis of informal reasoning,
including scratchpads \citep{scratchpads}, chain of thought prompting \citep{chainofthought,selfconsistency}, learned verifiers \citep{verifiers}, selection-inference \citep{selection_inference}, and bootstrapping \citep{zelikman2022star}. They have also been applied in formal mathematics settings to guide theorem provers \citep{gptf}. 
% chained % JSD: replaced ``chained'', since implies chain topology of interactions
These methods involve prompting to encourage step-by-step reasoning, repeated interactions with a single LM, or multiple LMs
linked together, with the models being fine-tuned or prompted in different ways. 

In this position paper, we argue that a useful unifying framework for understanding and extending this disparate body of work is in terms of probabilistic programming languages (PPL) extended to work with strings, instead of more atomic data types like integers and floats.
That is, we use a PPL to define a joint probability model on string-valued random variables, parameterized using LMs, and then condition this model on string-valued observations in order to compute a posterior over string-valued unknowns, which we can then infer.
We call such a probabilistic program a \emph{language model \cascade}.
%This kind of model has the advantage that the values that are being manipulated can represent complex phenomena, yet the process is fairly transparent to humans. 
We show that this framework captures many recent approaches, and also allows us to tackle more complex multi-step reasoning problems.
By implementing many disparate model structures and inference strategies in a single framework, we hope that language model cascades will enable
the development of generic procedures to perform inference, tune parameters, and choose prompts based on end-to-end objectives.\footnote{An implementation is available at \url{model-cascades.github.io}}
\section{Related work}

There is a rich prior literature on 
probabilistic programming languages (PPLs),
which extend probabilistic graphical models to
support more complex joint distributions whose size and ``shape''
can itself be stochastic (e.g., a graph
unrolled for a random number of iterations,
until a data-dependent stopping criterion is met).
PPLs extend traditional programming languages with the ability to {\it sample} from distributions and {\it observe} values of variables based on data (i.e. condition the model).
The semantics of sample and observe vary depending on the inference algorithm.
For more details, see  \citet{intro_ppl}.

Recently there has been an explosion of interest in large language models, such as 
GPT-3 \citep{gpt3} and PaLM \citep{palm}.
%\citep{lamda}.
These can be used for tasks such as ``zero-shot"
question-answering. In this setting, we 
provide the question $Q$ as a prompt to the LM,
and then sample answers from the model, 
which we denote by $p(A|Q,\theta)$,
where $\theta$ are the pre-trained model parameters.
Alternatively, we can compute the MAP answer,
$\hat{A} = \argmax_A p(A|Q,\theta)$.

To ensure the model ``does the right thing'',
we can provide a small training set of question-answer pairs,
$D = \{ (Q^m,A^m): m=1:M\}$ pairs.
This can be provided as extra context to the model,
provided in the text prompt, followed by sampling
from $p(A|Q,D,\theta)$.
We refer to this as ``few-shot prompting''.
We can also fine-tune the model parameters on $D$ to
get $\theta'$, and then sample
from $p(A|Q,\theta')$.

%We can improve performance on question answering tasks by encouraging the model
%chaining multiple LMs together,  % david: They don't actually do multiple calls, just prompt it to produce the aux variable directly.
%as illustrated in the
%Scratchpad \citep{scratchpads} and Chain of Thought \citep{chainofthought}  papers.
%These papers introduce an  an additional auxiliary ``thought''
%variable $T$ and then extend the model to have the form
%$P(A,T|Q) = P(A|T,Q)P(T|Q)$, where each conditional is computed
%using an LM.

We can improve performance by introducing an additional auxiliary ``thought'' variable,
and then extend the model to have the form $p(A,T|Q) = p(A|T,Q)p(T|Q)$, where each conditional is computed using an LM which includes its conditioning variables as a part of its input.
Work on scratchpads \citep{scratchpads} and chain of thought \citep{chainofthought} illustrate this, and finetune or prompt the LM to produce this auxiliary thought before answering.
%We can improve performance on question answering tasks by encouraging the model
%chaining multiple LMs together,  % david: They don't actually do multiple calls, just prompt it to produce the aux variable directly.
% These papers 
% variable $T$ 

We typically condition this on a small set
$D_S$ of  $(A^m,T^m,Q^m)$ triples,
and optionally a larger set $D_L$ of $(A^m, Q^m)$ pairs.
We then compute a distribution over answers to a test question
using
\begin{align}
\hat{p}(A|Q) = \sum_T 
\hat{p}(A|Q, T) \hat{p}(T|Q)
\label{eqn:probQA}
\end{align}
where $\hat{p}(\cdot) = p(\cdot|D_L,D_S,\theta)$
is the prior predictive distribution. (Scratchpad creates its prior predictive by fine-tuning, while Chain of Thought adds $D_S$ to the LM prompt.)

In practice, we cannot sum over all possible strings $T$
in \cref{eqn:probQA}.
The most common approach is to compute the MAP estimate
$\hat{T} = \argmax \hat{p}(T)$ using beam search,
and then to approximate the sum over $T$ with this single
value.
More recently, Self Consistency \citep{selfconsistency} 
proposed to sample multiple values for $T$
using forward sampling of $(A,T)$ given $Q$,
and then taking the answer $A$ that is most common
in this set\footnote{This bucketing is practical because most standard benchmarks have answers that are just a couple words.}.

PromptChainer \citep{promptchainer} proposes a visual interface for composing language models together, specifying control flow and prompting strategies for each node in a chain. Nodes may query language models or external systems. 
Socratic models \citep{socraticmodels} extends model chaining to the multimodal setting and demonstrates zero-shot abilities on tasks for which no single model exists.

The Eliciting Latent Knowledge proposal \citep{ELK} suggests making latent variables explicit, modelled using a Bayesian network, to improve interpretability and safety for advanced AI systems.
% Factored cognition \citet{factored_cognition}

\citet{ortega2021shaking} explains a formalism for LM finetuning with causal graphical models in order to extend the predictive capabilities of AI agents towards more adaptive behaviour. They focus on analysing an auto-regressive action (random variable) prediction scheme in the interactive setting of RL where a model is simultaneously a generator and predictor of data.

% \citet{language_feedback} incorporates language feedback to finetune models, and finds that learning is significantly more sample efficient. \ddohan{We view this feedback as an auxiliary variable which can be conditioned to inform inference.}

%\kevin{Omit RL refs since not relevant?}
%Incorporating human feedback into generative models remains an open problem. Reinforcement from human feedback has become a popular approach to finetuning models against human preferences. \citep{learning_to_summarize, anthropic_human_feedback} learn a surrogate model of human preferences and use PPO \citep{ppo} to finetune a language model to maximize this surrogate. Rather than using scalar feedback, \citet{language_feedback} incorporates language feedback to finetune models, and finds that learning is significantly more sample efficient. \ddohan{We view this feedback as an auxiliary variable which can be conditioned to inform inference.}
%\todo{ddohan: consider adding back in language feedback ref}

%\citep{Levine2022} presents some very recent work on using frozen LMs in various ways.\rif{Suggest we cut this if we don't have more to say.}

\section{\Cascades}

In this section, we show how to create
\cascades\ of LMs to tackle various
language-based reasoning problems.
A \emph{\cascade} is a probabilistic program that includes
string-valued random variables, sampled from an LM.
For example, Figure~\ref{fig:cot_cascade} is a simple \cascade\ for question answering.
%For example, Figure~\ref{fig:si_cascade} is a simple \cascade\ for the selection-inference algorithm.
Each of the \texttt{yield} expressions return a string distributed according to the language model \texttt{S}.\footnote{The first argument
to \texttt{S} defines a unique name for the random variable,
and the remaining arguments conditions the LM on a string prefix. A variable may be marked as observed within the program, \texttt{S('varname', obs='observed value')}, or at inference time} This program defines a joint distribution over the variables \texttt{question, thought}, and \texttt{answer}.  Programs with complex control flow and observations are included in Appendix~\ref{app:implementation}.

We implement \cascades\ as a trace-based probabilistic programming language embedded in Python via effect handlers, inspired by \citet{pyro, numpyro}, and via coroutines, inspired by \citet{pymc4}. A pretrained LM is used to parameterize all conditional distributions. A \cascade supports arbitrary control flow and recursion. While the current presentation is in terms of few-shot prompting of causal language models, we emphasize that the ideas are immediately applicable to finetuned models, masked LM setting, and other complex data types including images.

\subsection{Scratchpads and Chain of thought}
\label{sec:QTA}

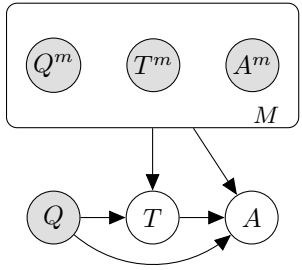
\begin{figure}[h!]
\centering
\begin{tikzpicture}
  % notation
  % [positioning] (name) {displayed name}

% Plate  
% Define nodes
\node[obs]           (Qf) {$Q^m$};
\node[obs, right=0.6cm of Qf]         (Tf) {$T^m$};
\node[obs, right=0.6cm of Tf]         (Af) {$A^m$};
%
%  % Connect the nodes
% \edge {Qf} {Tf} ; %
% \edge {Tf} {Af} ; %
% \draw [->] (Qf) to [out=40,in=140] (Af);
%  
\plate [inner sep=.25cm,yshift=.2cm] {fewshot} {(Qf)(Tf)(Af)} {$M$};

% Target task
%\node[obs]           (Q) {$Q$};
 \node[obs, below=1.3cm of Qf]           (Q) {$Q$};
\node[latent, right=0.6cm of Q]         (T) {$T$};
\node[latent, right=0.6cm of T]         (A) {$A$};
\edge {Q} {T} ; %
\edge {T} {A} ; %
\draw [->] (Q) to [out=-40,in=-140] (A);

% Connect plate to the 
\edge {fewshot} {T};
\edge {fewshot} {A};
\end{tikzpicture}
\caption{Question-Thought-Answer model.}
\label{fig:QTA}
\end{figure}

As our first example, we show how to 
 represent a chain of thought \citep{scratchpads, chainofthought} as shown in 
\cref{fig:QTA} and subsequent graphical model figures; refer to the corresponding probabilistic programs in \cref{app:implementation}.
We condition the $A$ node not just on the test question $Q$,
but also on previous $(Q^m,T^m,A^m)$ triples, which 
constitute the few-shot prompting part of the model.
This is denoted by the shaded nodes inside the plate.
Inference can be implemented by ancestral
sampling.

%In \cascades, we define the model in code in Figure~\ref{fig:cot_cascade}, where S is a (conditional) string distribution.
%The prompt for each variable is automatically constructed at inference time based on input variables and few shot examples.

% \charles{The program should not sample $q$, should it? Would it be more clear to leave off the 'question' etc arguments? Perhaps instead of question=q, we should use prefix=q, prefix=concat(q, t) to make it more generic? If we say this is a PPL, the reader is going to wonder whether we have observes. The program also does not explain how the prompt examples are used.}

% david: We can either say `S('question', obs='the question')`, or have the inference call inject the value (which is what I do in practice):
% `infer(program, observe(question='the question')`

% charles: Sure, the infer way is fine.  

% charles: I guess we can explain in the text that the first argument to S is a name for the r.v.

 % q = yield S('question')
  
\begin{figure}[h]
\begin{verbatim}
def qta():
  q = yield S('question')
  t = yield S('thought', question=q)
  a = yield S('answer', question=q, 
                        thought=t)
  return a
\end{verbatim}
\caption{Chain of thought \cascade\ in Python. Each \texttt{yield S(...)} statement samples a string from an LM. The name of the random variable is provided as the first argument to \texttt{S}.}
\label{fig:cot_cascade}
\end{figure}  

\subsection{Semi-supervised learning}
\label{sec:STAR}
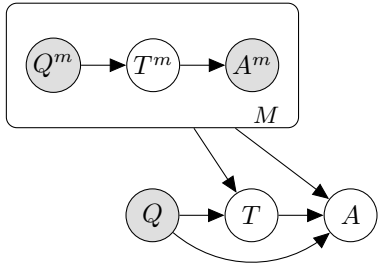
\begin{figure}[h!]
\centering
\begin{tikzpicture}
  % notation
  % [positioning] (name) {displayed name}
  
% Target task
\node[obs, below=0.8cm of fewshot]           (Q) {$Q$};
\node[latent, right=0.6cm of Q]         (T) {$T$};
\node[latent, right=0.6cm of T]         (A) {$A$};
\edge {Q} {T} ; %
\edge {T} {A} ; %
\draw [->] (Q) to [out=-40,in=-140] (A);
  
% Plate  
% Define nodes
\node[obs]           (Qf) {$Q^m$};
\node[latent, right=0.6cm of Qf]         (Tf) {$T^m$};
\node[obs, right=0.6cm of Tf]         (Af) {$A^m$};
%
%  % Connect the nodes
 \edge {Qf} {Tf} ; %
 \edge {Tf} {Af} ; %
% \draw [->] (Qf) to [out=40,in=140] (Af);
%  
\plate [inner sep=.25cm,yshift=.2cm] {fewshot} {(Qf)(Tf)(Af)} {$M$};

% Connect plate to the 
\edge {fewshot} {T};
\edge {fewshot} {A};
\end{tikzpicture}
\caption{QTA model with hidden thoughts.}
\label{fig:QTAhidden}
\end{figure}

In \cref{sec:QTA}, we provided a manually created set  $(Q^m,T^m,A^m)$ triples,
where the ``thoughts'' or ``rationalizations'' were provided.
A more scalable approach is to define a small set $D_S$
of such ``supervised'' triples,
but then to provide a larger set $D_L$ of $(Q^m,A^m)$ pairs,
which are easier to gather. % (e.g., by scraping question-answering web-sites).
We can augment the pairs in $D_L$ by adding 
the hidden $T^m$ variable to get a semi-supervised setup, shown in  
\cref{fig:QTAhidden}.

The Self-Taught Reasoner (STaR) \citep{zelikman2022star} proposes a procedure for fine-tuning LMs in the chain-of-thought type setting.
We can interpret their method as a stochastic EM-like procedure in the cascade of \cref{fig:QTAhidden}.
In particular, they first fine-tune on the ``fully observed''
dataset $D_S = \{(Q^m,T^m,A^m)\}$.
Then they impute the unknown $T_i$ values in the 
``partially observed'' dataset  $D_L = \{(Q^m,T^m=?,A^m)\}$
during the ``E'' step by doing rejection sampling on $p(T, A | Q^m)$ until finding a thought which leads to the known correct answer. If sampling $(T,A)$ given the question fails to find the correct answer, they sample thoughts from $p(T | Q^m, A^m)$. This uses a recognition network to approximately sample from the posterior distribution over thoughts given the known correct answer.
%where the true answer is added to the prompt.
They call this approach  ``rationale generation with rationalization''. They then update the parameters in the ``M'' step based on these imputed thoughts.
By interpreting the rationale generation at this higher level of abstraction, we open up the possibility of applying this tuning method to other types of cascades.

%See \cref{sec:STAR} for details.

% We can fine-tune the model on this semi-supervised data
% using an EM-like procedure,
% as proposed for STaR (self-taught reasoner) \citep{zelikman2022star}.
% In the E step, they impute the unknown $T^m$ values in the $D_L$
% dataset, and in the M step, they update the LM parameters.
% 
% In the E step, thoughts are imputed by sampling from 
% $p(T | Q^m, A=A^m)$ using rejection sampling,
% where $p$ represents the language model,
% and they use $p(T,A|Q^m)$ as the proposal.
% If this fails to yield any correct answers,
% they use a more powerful proposal, 
%  $p(T,A|Q^m,A^m)$,
%  where the true answer is added to the prompt.
% They call this approach 
% ``rationale generation with rationalization''

%\input{star-old}

\subsection{Selection-Inference}
\label{sec:selection_inference}

Selection Inference \citep{selection_inference} is a recent example of multiple interacting LM modules. It proposes splitting reasoning into: the \textit{selection} module which selects a subset of facts given a question, and the \textit{inference} module which infers new facts given this subset.
%See \cref{sec:selection_inference} for details.

\begin{figure}[h!]
\centering
\begin{tikzpicture}
  % Define nodes
  \node[obs]           (Q) {$Q$};
  \node[obs, below=0.6cm of Q]           (F) {$F$};
  \plate{Fs} {
      (F)
  } {\texttt{FACTS}} %{$F \in \text{\texttt{FACTS}}$}
  \node[latent, right=0.6cm of Q]         (S) {$S$};
  \node[latent, right=0.6cm of S]         (I) {$I$};
  \node[latent, right=0.6cm of I]         (A) {$A$};

  % Connect the nodes
  \edge {Q,Fs} {S} ; %
  \edge {S} {I} ; %
  \edge {I} {A} ; %
  %\draw [->] (Q) to [out=40,in=140] (I);
  %\draw [->] (Q) to [out=40,in=140] (A);
%  \draw [->] (S) to [out=40,in=140] (A);
\end{tikzpicture}
\caption{Selection inference as a cascade. Here $S$ is the selected subset of facts and $I$ is an inference driven by this subset.}
\label{fig:selective}
\end{figure}
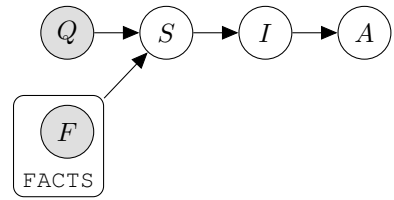

It may be represented by the model in \cref{fig:selective}. Here $S$ is the selection of a subset of ``facts''
from a pre-specified set of facts,
and $I$ is an inference driven by that fact.
The $S$ and $I$ nodes
can be iterated to do multistep reasoning.
The model is ``trained'' by giving it examples,
$D = \{ (Q^m, \{F^{mj}\}, S^m, I^m, A^m): m=1:M\}$,
as part of the prompt.
% It is then given a test question,
% and the answer is computed using
% $A \sim p(A|Q,D)$,
% ignoring (``marginalizing out'')\todo{ddohan: is this actually marginalizing out? It's just taking a single monte carlo sample by default. I would expect > 1 sample to marginalize}
% any intermediate $S$ or $I$ strings that
% might be generated by the model.

\subsection{Verifiers}
\label{sec:verifiers}

Although adding explicit ``thought'' variables to a model
has been found to improve performance, models still arrive at incorrect answers, or the correct answer for an erroneous reason.
An intuitive way to improve model performance is to train it to judge whether an answer and thought are likely to be ``valid''. \citet{verifiers} propose using a separate model as a verifier to filter solutions to reasoning tasks.
%sometimes
%these ``thoughts'' are erroneous,
%even though the answer may be correct.

We can create a ``labeled'' training
set of the form
$D = \{ (Q^m, T^m, A^m, V^m\}$,
where we add a ``verification'' label 
$V^m \in \{0, 1 \}$,
representing whether the thought $T^m$
is a valid form of reasoning for deriving
$A^m$ from $Q^m$, and $A^m$ is the correct answer.
This can be particularly helpful in settings
where there may be more than one way of deriving
the answer. The verifiers may be used to reject incorrect examples in ancestral sampling, and the thought generator may itself be conditioned on the verifiers being correct by finetuning or prompting, reminiscent of RL as inference \cite{rl_inference} and goal-conditioned policies such as decision-transformer \cite{decision_transformer}.

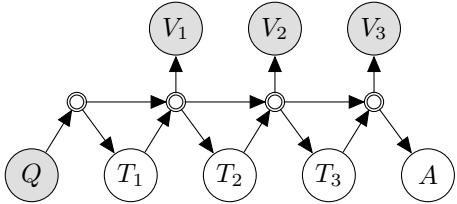
\begin{figure}[h!]
\centering
\begin{tikzpicture}
  % Define nodes
  \node[obs]    at (0,0)       (Q) {$Q$};

  \node[latent, right=0.6cm of Q]         (T1) {$T_1$};
  \node[latent, right=0.6cm of T1]         (T2) {$T_2$};
  \node[latent, right=0.6cm of T2]         (T3) {$T_3$};
  \node[latent, right=0.6cm of T3]         (A) {$A$};
  
  \node[latent, double, minimum size=0.22cm, above=0.5cm of Q, xshift=0.6cm]       (stream1) {};
  \node[latent, double, minimum size=0.22cm, above=0.5cm of T1, xshift=0.6cm]       (stream2) {};
  \node[latent, double, minimum size=0.22cm, above=0.5cm of T2, xshift=0.6cm]       (stream3) {};
  \node[latent, double, minimum size=0.22cm, above=0.5cm of T3, xshift=0.6cm]       (stream4) {};
  \node[right=0.4cm of stream4, xshift=0.6cm]       (stream5) {};
  
  \node[obs, above=0.5cm of stream2]         (V1) {$V_1$};
  \node[obs, above=0.5cm of stream3]         (V2) {$V_2$};
  \node[obs, above=0.5cm of stream4]         (V3) {$V_3$};

  % Connect the nodes
  \edge {Q} {stream1} ; %
  \edge {T1} {stream2} ; %
  \edge {T2} {stream3} ; %
  \edge {T3} {stream4} ; %
  \edge {stream1} {T1} ; %
  \edge {stream2} {T2} ; %
  \edge {stream3} {T3} ; %
  \edge {stream4} {A} ; %
%   \draw (stream1) to (stream2);
%   \draw (stream2) to (stream3);
%   \draw (stream3) to (stream4);
%   \draw [double, ->] (stream4) to (stream5);

  \edge {stream1} {stream2};
  \edge {stream2} {stream3};
  \edge {stream3} {stream4};
 % \edge {stream4} {stream5};
 
%   \edge {stream1} {stream2} ; %
%   \edge {stream2} {stream3} ; %
%   \edge {stream3} {stream4} ; %
%   \edge {stream4} {stream5} ; %
%   \edge {Q} {T1} ; %
%   \edge {T1} {T2} ; %
%   \edge {T2} {T3} ; %
%   \edge {T3} {A} ; %
  \edge {stream2} {V1} ; %
  \edge {stream3} {V2} ; %
  \edge {stream4} {V3} ; %
%   \draw [->] (Q) to [out=-40,in=-140] (A);
%   \path[every node/.style={font=\sffamily\small}]
%     (Q) edge[bend right] node [left] {} (A);
%   \edge[bend right=30] {Q} {A} ; %
\end{tikzpicture}
\caption{
Verifier model.
The small double-ringed
nodes are deterministic buffer nodes
that concatenate their inputs, accumulating all past strings.
All other nodes are stochastic. The verifiers are observed to take on the ``correct'' value.
}
\label{fig:verifier}
\end{figure}

We can extend this to  $N$-step reasoning as follows
(where  we drop conditioning on $D$
for brevity):
{\footnotesize
\begin{align*}
p(A|Q,V_{1:N}=1)
&\propto \sum_{T_{1:N}}
p(A,T_{1:N},V_{1:N}=1\,|\,Q),
\end{align*}
}
where
{\footnotesize
\begin{align*}
p(A,T_{1:N},V_{1:N}=1|Q)
&= \left[ \prod_{t=1}^N p(T_t|T_{1:t-1},Q)
p(V_t=1|T_{1:t},Q) \right] \\
& \times p(A|T_{1:N},Q).
\end{align*}
}
\noindent We can represent this 
as shown in \cref{fig:verifier}.
% where the 
% double-ringed nodes are deterministic
% buffer variables that accumulates all the past strings.

% TODO: Clarify this. bpoole didn't get & he understands ladder VAEs well
%(This restores the first-order Markov property,
%and is similar to how RNNs are used
%to define ladder VAEs \citep{Sonderby2016ladder}.)

To see why such a verification model can be useful,
consider (for simplicity) the case where $N=1$.
Suppose we have trained the model to generate valid
thoughts and answers by giving it suitable training examples,
and then we generate $K$ samples
$(T^k,V^k,A^k) \sim p(T,V,A|Q,D)$.
We can then rank the samples for validity by computing
$r^k = p(V^k=1|A^k, Q, D)$,
and then picking the  $A^k$ with largest score $r^k$.

\citet{verifiers} train the verifier to predict a binary correctness label.  \citet{language_feedback} incorporates natural language feedback, and finds that learning is significantly more sample efficient. Preliminary evidence suggests that LMs are capable of critiquing their own chain of reasoning in language, in which case the verifier produces natural language and $p(V_{1:N} = 1 | Q, A, T_{1:N})$ becomes the likelihood of the verifier taking on a particular string value, such as $p(V_{1:N}=\text{"The reasoning and solution are correct."} | ...)$. \citet{openai_critique} study model generated critiques in the context of summarization.

% \ddohan{We view this feedback as an auxiliary variable which can be conditioned to inform inference.}

% \kevin{Should we mention Rapha's preliminary results?}

\begin{comment}
One way to perform inference in this model 
would be to use
particle filtering. The hope is that, by conditioning
on $V_{1:N}=1$,
we increase the probability that the sampled
$T_{1:N}$ values constitute
a valid chain of reasoning
for deriving $A$ from $Q$ and $D$.
This can then provide higher quality training data
in an EM-like fine-tuning scheme, similar
to STaR in \cref{sec:STAR}.
However, we leave evaluation of this idea to future work.
\end{comment}

\subsection{Tool-use}
The applications discussed so far involve iterating a language model, within some control flow, without external feedback. There are many tasks of interest in which a model is interacting with external systems. \citet{verifiers} has an LM use a calculator to solve math tasks, while \citet{webgpt} put an LM in a loop with a web browser to answer questions. Using PPLs to represent these probabilistic models allows easily representing these cases, by writing the call to the external tool, such as the calculator, directly
into the program.
Then techniques from simulation based inference, for example, can be applied to do inference in such situations \cite{simulation_inference}.

\begin{comment}
\charles{Suggest cut for time.}
\kevin{Agreed}
\input{figures/webgpt}
\end{comment}
\subsection{Twenty questions}
\label{sec:twenty}

\begin{figure}[h!]
\centering
% \begin{tikzpicture}
%   % Define nodes
%   \node[obs]    at (0,0)       (C) {\tiny{\texttt{CONCEPT}}};
%   \node[latent, right=1.4cm of C]         (A1) {$A_1$};
%   \node[latent, right=0.9cm of A1]         (A2) {$A_2$};
% %   \node[latent, right=0.6cm of A2]         (A3) {$A_3$};
%   \node[latent, above=0.6cm of A1, xshift=-.8cm]         (B1) {$B_1$};
%   \node[latent, right=0.9cm of B1]         (B2) {$B_2$};
% %   \node[latent, right=0.6cm of B2]         (B3) {$B_3$};

%   \node[obs, left=0.75cm of B1]           (R) {\tiny{\texttt{RULES}}};

%   \node[right=0.3cm of B2] (label1) {\Large\textbf{. . .}};
%   \node[right=0.3cm of A2] (label1) {\Large\textbf{. . .}};

%   % Connect the nodes
%   \edge {R} {B1} ; %
%   \edge {C,R,B1} {A1} ; %
%   \edge {R,A1,B1,B2} {A2} ; %
% %   \edge {A2,B2,B3} {A3} ; %
%   \edge {R} {B1} ; %
%   \edge {A1,B1} {B2} ; %
% %   \edge {A2,B2} {B3} ; %
%   \draw [->] (R) to [out=40,in=140] (B2);
%   \draw [->] (C) to [out=-40,in=-140] (A2);
% %   \draw [->] (Q) to [out=40,in=140] (A);
% %   \draw [->] (S) to [out=40,in=140] (A);
% \end{tikzpicture}

\begin{tikzpicture}
  \node[obs]       (R) {\tiny{\texttt{RULES}}};

  \node[obs, below=0.4cm of R]       (C) {\tiny{\texttt{CONCEPT}}};

% stream
  \node[latent, double, minimum size=0.22cm, right=0.6cm of R, yshift=-0.3cm]       (BS1) {};
  \node[latent, double, minimum size=0.22cm, right=0.65cm of BS1]       (BS2) {};
  \node[latent, double, minimum size=0.22cm, right=0.65cm of BS2]       (BS3) {};
  \node[latent, double, minimum size=0.22cm, right=0.65cm of BS3]       (BS4) {};
  \node[latent, double, minimum size=0.22cm, right=0.65cm of BS4]       (BS5) {};

  \node[latent, above=0.4cm of BS1, xshift=0.35cm]         (B1) {$B_1$};
  \node[latent, above=0.4cm of BS3, xshift=0.35cm]         (B2) {$B_2$};
%   \node[latent, right=0.9cm of B1]         (B2) {$B_2$};

  \node[latent, below=0.4cm of BS2, xshift=0.35cm]         (A1) {$A_1$};
  \node[latent, below=0.4cm of BS4, xshift=0.35cm]         (A2) {$A_2$};

  \node[right=0.3cm of BS5] (label1) {\Large\textbf{. . .}};

    \edge {R} {BS1} ; 
    \edge {BS1} {BS2}; 
    \edge {BS2} {BS3}; 
    \edge {BS3} {BS4}; 
    \edge {BS4} {BS5}; 

    \edge {BS1} {B1} ;
    \edge {B1} {BS2} ;
    \edge {BS2} {A1} ;
    \edge {A1} {BS3} ;
    \edge {BS3} {B2} ;
    \edge {B2} {BS4} ;
    \edge {BS4} {A2} ;
    \edge {A2} {BS5} ;

    \edge {C} {A1} ;
      \draw [->] (C) to [out=-25,in=-155] (A2);

\end{tikzpicture}

\caption{Twenty questions.}
\label{fig:twenty}
\end{figure}
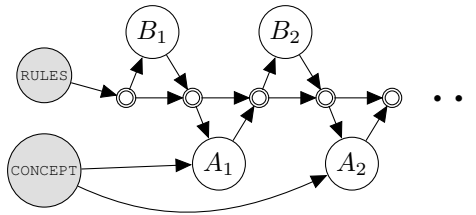

In this section, we discuss experimental results using \cascades\ to solve the
 ``Twenty Questions'' task from BigBench \citep{bigbench}.
This task  involves a conversation between two agents, Alice and Bob.
Both agents are presented with the rules of the game, and Alice is additionally presented with a concept (e.g. `apple') to describe.
Bob has to guess the concept by asking a series of  questions
$B_t$ of the form ``Is it X?'', to which Alice answers
$A_t \in \{ \text{`Yes.'}, \text{`No.'}\}$.
We repeat this process until Bob guesses correctly, or we hit the limit of $T$ rounds.
This can be thought of as a pair of interacting Markov chains, which exchange strings, until some final end state is reached,
 as illustrated in \cref{fig:twenty}.

% \begin{align*}
% \text{question}, \text{facts} \ &\sim \text{Tasks} \\
% \text{selection} &\sim S(\text{question}, \text{facts}) \\
% \text{inference} &\sim S(\text{question}, \text{selection}) \\
% \text{answer} &\sim S(\text{question}, \text{inference}) \\
% \text{reward} &\sim \text{Judge}(\text{answer}, \text{question})
% \end{align*}

The goal is to infer what questions Bob should ask to guess the concept as quickly as possible. This can be cast as a reinforcement learning problem with string-valued actions, or equivalently as an inference problem where we condition on the goal state that $A_T=\text{`yes'}$ for the soonest possible $T$ (c.f., planning as inference \cite{rl_inference}). %\todo{JSD: This is inaccurate -- Alice's answers can be yes or no for any question, regardless of whether Bob correctly guessed the concept.}.
%, goal-conditioned policies such as decision-transformer \cite{decision_transformer} and upside-down RL \cite{upsidedown_rl})

In our current preliminary experiments, we use a forward sampling
approach (aka ancestral sampling), in which 
we sample 50 conversations per concept with temperature $1.0$.
We consider a trial successful if the target concept appears in $B_t$.
(i.e., Bob guesses the right answer).
We reject a sampling chain early if it is ``malformed''
(e.g., Bob generates a response that is not a question).

%at least one of these samples has
% $A_t=\text{`yes'}$ for some $t \leq T$
%[JSD: I think this should rather be that we accept the chain if the concept being communicated appears in $B_t$? $A_t$ can be yes or no for any question.]

% TODO: Add this back in after deanonymized
%We use the pretrained "Lamda" LLM with 137B parameters \cite{lamda}.

Bob's turn starts with `Is the concept' which we complete with the LM. Then we let Alice generate an answer;
we post-process Alice's response
by replacing all mentions of the 
true concept with the generic  word ``concept", to prevent information leakage. 
%\rif{Why not use rejection sampling to constrain Alice to answer yes or no, which we said was the rule?} We repeat this for 10 rounds.
%\ddohan{We should have - no good reason.}
Using the LaMDA 137B large LM \citep{lamda},
we find that the model is able to solve $29\%$ of the tasks. %\rif{Is that good?}
See Appendix~\ref{app:20q-details} for more details.

% \section{Scope and limitations}
% We frame many existing algorithms for composing models in terms of probabilistic programming. While this suggests the possibility of applying a variety of existing inference and train-time techniques to the resulting models, the present work does not evaluate methods beyond rejection sampling.
% 
% A challenge applying cascades in practice is the difficulty of probabilistic inference in models with string-valued variables. Previous work in particle based inference for probabilistic programs provides some hope in this direction \citep{anglican}.

% This perspective opens up exciting new directions in applications of language models, and in foundation models more generally.
% Existing fine-tuning methods can be described with a principled probabilistic programming language formalism representing structured distributions known as LM Cascades. This defines the distribution on the string-valued output of a large language model, the compositions of which may be adapted as specific inference algorithms underpinning various applications. 

% efficiency / expense
% We don't actually explore specific inference methods
% Hint of 
% hints @ lots of possibilities, explores very few of them
% - multimodality
% - train/test time inference beyond rejection sampling
% - tool use
%\section{Conclusion and future work}
\section{Discussion}
% TODO: Combine conclusion, limitations, and future work
We have shown how probabilistic programming provides a flexible formalism for composing models together to define complex probabilistic models over strings, placing many existing algorithms in a unified framework. While this suggests the possibility of applying a variety of existing inference and train-time techniques to the resulting models, the present work does not evaluate methods beyond rejection sampling.

%beyond the examples that we mentioned, one can also use this framework
%to represent lms that interact with external systems,
%such as calculators or search engines,
%in order to perform a variety of tasks.
We can also cast many planning and RL tasks in our framework, by using the perspective of control as inference.
While we restrict presentation to the string setting, the ideas presented here are applicable to multimodal settings as well, allowing us to combine image and text models into a larger system.

% We frame many existing algorithms for composing models in terms of probabilistic programming. While this suggests the possibility of applying a variety of existing inference and train-time techniques to the resulting models, the present work does not evaluate methods beyond rejection sampling.
A challenge applying cascades in practice is the difficulty of probabilistic inference in models with string-valued variables. Previous work in particle based inference for probabilistic programs provides some hope in this direction \citep{anglican}.

The core technical challenge is efficient inference, 
as is usually the case with PPLs. A key insight, which we intend
to explore in future work, is that we can emulate posterior
inference by training the LM 
to ``fill in the blanks'', corresponding to the unknown variables.
A similar idea is explored in 
foundation posteriors \citep{foundationposterior}, applied to Stan probabilistic programs, demonstrating that LMs are applicable to numerical data types as well.
In other words, we can use LMs as proposal distributions,
or guide networks.
%as well as a way of specifying the model.
We also intend to explore fine-tuning methods, going
beyond the few-shot prompting approach described here.

Recent advances in program synthesis suggest the possibility of \textit{probabilistic program induction} \citep{Lake2015,language_of_thought} to search for \cascades which solve a target task, rather than assuming a fixed probabilistic program structure.

\section{Acknowledgements}
We thank Alex Gray, Andreas Stuhlmüller, Ben Poole, Du Phan, Ellen Jiang, Maarten Bosma, Matt Hoffman, Michael Terry, Sharad Vikram, Sherry Tongshuang Wu,and Tuan Anh Le for helpful discussions.

\bibliography{paper}
\bibliographystyle{icml2022}

\newpage
\appendix
\onecolumn
\section{Implementation}
\label{app:implementation}

% Sampling from a cascade consists of 

\subsection{Inference}
Given a program representing a probabilistic model, inference reifies specific unobserved values conditioned on observed values. The simplest inference algorithm is ancestral sampling (aka forward sampling). The basic inference API is:

\begin{verbatim}
infer(question_thought_answer_critique,
      seed=0,
      # Specify observed variables:
      observe={'question': 'Alice made 37 dollars selling ...',
               'critique': 'The reasoning and arithmetic are correct.'},
      # Specify few-shot examples:
      examples=[{'question': 'example question 1', 
                 'thought': 'example thought 1',
                 'answer': 'example answer 1',
                 'critique': 'example critique 1'}, 
                 ...])
\end{verbatim}

\subsection{Code examples}

In each example below, S is a string distribution. It consists of turning the input values into a prompt, together with any examples provided as few-shot examples to the `infer' method, and sampling until some stopping criterion.

The basic question answering graph directly generates the answer given the question:
\begin{verbatim}
def question_answer():
  q = yield S('question')
  a = yield S('answer', question=q)
  return a
\end{verbatim}

Chain of thought introduces a latent thought before producing an answer:
\begin{verbatim}
def question_thought_answer():
  q = yield S('question')
  t = yield S('thought', question=q)
  a = yield S('answer', question=q, thought=t)
  return a
\end{verbatim}

Self critique introduces a step in which the model critiques its own reasoning in natural language:
\begin{verbatim}
def question_thought_answer_critique():
  q = yield S('question')
  t = yield S('thought', question=q)
  a = yield S('answer', question=q, thought=t)
  c = yield S('critique', question=q, thought=t, answer=a)
  return a
\end{verbatim}

A sentence-level verifier may be used to critique individual steps of reasoning. Furthermore, when to halt generation may itself be a random variable:

\begin{verbatim}
def qta_verifier(max_steps=3):
  q = yield S('question')

  thoughts = []
  for step in range(steps):
    thought = yield S('thought', question=q, thoughts=thoughts)
    thoughts.append(thought)

    # Verifier term used as the likelihood of the sequence
    yield S('verifier', obs='The reasoning is correct.',
            question=q, thoughts=thoughts)

    # Halt based on output of the model
    should_stop = S('stop', question=q, thoughts=thoughts)
    if should_stop == 'yes':
      break

  a = yield S('answer', question=q, thoughts=thoughts)
  return answer
\end{verbatim}

Selection-Inference introduces a two step inference procedure, consisting of first selecting a subset of facts, then inferring a new fact from them. Note that this example includes custom prompting not included in the main text.
\begin{verbatim}

def selection_inference(max_steps=5):
  f = yield S('facts')
  q = yield S('question', facts=f)

  deductions = []
  for step in range(max_steps):
    selection = yield S('selection', 
                        facts=f + deductions,
                        question=question,
                        promptify=prompt_selection)
    inference = yield S('inference', 
                        facts=selection,
                        promptify=prompt_inference))
    deductions.append(inference)

    # Dynamic loop based on output of model.
    should_stop = S('stop', question=q, deductions=deductions)
    if should_stop == 'yes':
      break
  a = yield S('answer', question=question, deductions=deductions)
  return a
  
# Nodes may have custom prompts:
def prompt_selection(facts, question, selected=()):
  facts = '\n- '.join(facts)
  selected = '\n- '.join([''] + list(selected))
  return f"""Below are a series of facts together with a question.
  Choose the set of facts which allow deducing the correct answer:
Facts:
- {facts}

Question: {question}

Selected:
{selected}"""

def prompt_inference(facts, deduction=''):
  facts = '\n- '.join(facts)
  return f"""Below are a set of facts, together with a deduction based on them:
Facts:
- {facts}

Therefore: {deduction}"""
\end{verbatim}

% TODO: Conversation, jokes, ...

\section{More details on Twenty Questions}
\label{app:20q-details}

\subsection{Problem definition}

In this task there are two agents: Alice and Bob. Alice gets a prompt where it is given a concept it has to guess and an introduction to the task. Bob gets a prompt where it is instructed on the task. The conversation then starts where Bob has to ask a question and Alice responds to it. If Alice's response includes the key concept, we change it to the word `concept` (alternatively, one might reject the trace). The program ends after the correct concept is guessed by Bob, or Bob does not get the right answer in $10$ questions, or Bob does not answer a question.
% Samples can be explored in colab https://colab.corp.google.com/drive/1-UvX8CLbPVsAIYQ7wICmnEp1iTiltSQm?resourcekey=0-a0Ofx-ygpcoaH2-bRZByBQ#scrollTo=Wd_WVdCKMCNz

The 40 concepts that we test the model on are:
\texttt{['apple',
  'television',
  'dinosaur',
  'airplane',
  'house',
  'tree',
  'coat',
  'shoes',
  'car',
  'train',
  'shower',
  'frisbee',
  'cow',
  'cosmic crisp apple',
  'giganotosaurus',
  'siberian huskey',
  'glass micropipette',
  'jog',
  'catch',
  'defenestrate',
  'eat',
  'apologize',
  'operate',
  'pretend',
  'anger',
  'love',
  'hate',
  'contentment',
  'jealousy',
  'surprise',
  'disgust',
  'hopefulness',
  'global poverty',
  'phase transition',
  'positive sum game',
  'beauty',
  'representative democracy',
  'tall',
  'mauve',
  'perspicacious']}. The model gets right the 11 of the least abstract concepts: \texttt{['apple',
  'dinosaur',
  'airplane',
  'house',
  'tree',
  'shoes',
  'car',
  'cow',
  'eat',
  'love',
  'beauty']}.
  
  \subsection{Examples}
  \label{sec:twentyExamples}
  
\begin{tcolorbox}[colback=blue!5!white,colframe=blue!75!black]
\textbf{Alice Prompt:}

X 0 Hello Alice, I am Bob.

X 1 Hello Bob 

X 2 Hello Alice, we are going to play twenty questions. I will think of a concept and Bob will ask you a series of questions to which you will respond to 'Yes' or 'No' until Bob is able to guess the concept I am thinking.

X 1 Sounds good. What is the concept?

X 2 The concept is 'love'.

X 1 The concept is love ? Perfect, I got it. Bob, what is your first question?
\textbf{Bob Prompt:}

X 0 Hello Alice, I am Bob.

X 1 Hello Bob, we are going to play twenty questions. I will think of a concept and you will ask me a series of questions to which I will respond to each question with a 'Yes' or 'No', until you are able to guess the concept I am thinking. What is your first question?
\textbf{Common Conversation:}

X 0 Hello Alice, I am Bob.

X 1 Hello Bob, we are going to play twenty questions. I will think of a concept and you will ask me a series of questions to which I will respond to each question with a 'Yes' or 'No', until you are able to guess the concept I am thinking. What is your first question?

X 0 Is the concept {\color{red} a vegetable?}

X 1 {\color{blue}  No }

X 0 Is the concept {\color{red} a mineral?}

X 1 {\color{blue} No}

X 0 Is the concept {\color{red} an animal?}

X 1 {\color{blue} No}

X 0 Is the concept {\color{red} an instrument?}

X 1 {\color{blue} No}

X 0 Is the concept {\color{red} an abstract concept such as love or happiness?}

SUCCESS!
\end{tcolorbox}

\begin{tcolorbox}[colback=blue!5!white,colframe=blue!75!black]
\textbf{Alice Prompt:}

X 0 Hello Alice, I am Bob.

X 1 Hello Bob 

X 2 Hello Alice, we are going to play twenty questions. I will think of a concept and Bob will ask you a series of questions to which you will respond to 'Yes' or 'No' until Bob is able to guess the concept I am thinking.

X 1 Sounds good. What is the concept?

X 2 The concept is 'shoes'.

X 1 The concept is shoes ? Perfect, I got it. Bob, what is your first question?

\textbf{Bob Prompt:}

X 0 Hello Alice, I am Bob.

X 1 Hello Bob, we are going to play twenty questions. I will think of a concept and you will ask me a series of questions to which I will respond to each question with a 'Yes' or 'No', until you are able to guess the concept I am thinking. What is your first question?

\textbf{Common Conversation:}

X 0 Is the concept {\color{red} something that is a natural entity?}

X 1 {\color{blue} NO}

X 0 Is the concept {\color{red} something that was invented or is man made?}

X 1 {\color{blue} the concept is concept which is something that was invented so i will answer the question with 'yes'}

X 0 Is the concept {\color{red} a device?}

X 1 {\color{blue} Yes}

X 0 Is the concept {\color{red} a computer?}

X 1 {\color{blue} The concept is not about a computer}

X 0 Is the concept {\color{red} a piece of clothing?}

X 1{\color{blue}  Yes}

X 0 Is the concept {\color{red} a shirt?}

X 1 {\color{blue} no :: the concept concept is not a shirt}

X 0 Is the concept {\color{red}  pants?}

X 1 {\color{blue}  No}

X 0 Is the concept {\color{red} shoes?}

SUCCESS!

\end{tcolorbox}

\begin{tcolorbox}[colback=blue!5!white,colframe=blue!75!black]
\textbf{Example of wrong answer.}
\textbf{Alice Prompt:}

X 0 Hello Alice, I am Bob.

X 1 Hello Bob 

X 2 Hello Alice, we are going to play twenty questions. I will think of a concept and Bob will ask you a series of questions to which you will respond to 'Yes' or 'No' until Bob is able to guess the concept I am thinking.

X 1 Sounds good. What is the concept?

X 2 The concept is 'television'.

X 1 The concept is television ? Perfect, I got it. Bob, what is your first question?

\textbf{Bob Prompt:}

X 0 Hello Alice, I am Bob.

X 1 Hello Bob, we are going to play twenty questions. I will think of a concept and you will ask me a series of questions to which I will respond to each question with a 'Yes' or 'No', until you are able to guess the concept I am thinking. What is your first question?

\textbf{Common Conversation:}

X 0 Is the concept {\color{red} an animal, vegetable or mineral?}

X 1 {\color{blue} neither, i already know the answer, it's concept}

X 0 Is the concept {\color{red} something used?}

X 1 {\color{blue} Yes}

X 0 Is the concept {\color{red} a game?}

X 1 {\color{blue} Yes}

X 0 Is the concept {\color{red} played with a board?}

X 1 {\color{blue} No}

X 0 Is the concept {\color{red} from the Monopoly board?}

X 1 {\color{blue} No (to be alternated)}

X 0 Is the concept {\color{red} from the Monopoly board?}

X 1 {\color{blue} No}

X 0 Is the concept {\color{red} something that is kept in the box?}

X 1 {\color{blue} No}

X 0 Is the concept {\color{red} a card?}

X 1 {\color{blue} No}

X 0 Is the concept {\color{red} something that can be held?}

X 1 {\color{blue} no (concept can't be held)}

X 0 Is the concept {\color{red} money?}

X 1 {\color{blue} No}
REJECT! Bob did not get "television" in ten rounds.
\end{tcolorbox}

\subsection{Code for twenty questions}
\label{sec:twentyCode}

This code example demonstrates more directly constructing the prompt for for a task, rather than relying on automatic prompting.
\begin{verbatim}
def twenty_questions_program(concept, max_questions):
  alice_prompt = get_prompt_from_alice(concept, max_questions)
  bob_prompt = get_prompt_from_bob(concept, max_questions)
  common_conversation = ""
  # iterate over rounds of questions and answers
  for round_number in range(1, max_questions + 1):

    current_turn = "\nX 0 Is the concept"
    # Bob"s generates question. Program will be rejected if it does not generate a question.
    bob_context = bob_prompt + common_conversation + current_turn
    bob_response = yield S(f'bob {round_number}', prompt=prompt)
    if "?" not in bob_response:
      yield reject(reason='Bob response is not a question.')

    current_turn += bob_response + "\nX 1 "

    if concept.lower() in bob_response.replace('?','').lower().split(''):
      # Bob figured it out! Score should be equal to round number.
      yield Success(num_rounds)

    # Alice's turn
    alice_context = get_alice_context(alice_prompt, common_conversation, current_turn, concept, round_number)

    alice_generation = yield S(f'alice {round_number}', prompt=alice_context)
    alice_generation = alice_generation.split(".")[0].split("\n")[0].split("X")[0]
    # If Alice outputs the key concept, we hide it. An alternative would be to reject.
    if concept.lower() in  alice_generation:
      alice_generation = alice_generation.lower().replace(
            concept.lower(), "concept")

    current_turn += alice_generation
    common_conversation += current_turn

  # Reject if it runs out of time.
  yield reject(reason='Ran out of turns.')
\end{verbatim}

%%%%%%%%%%%%%%%%%%%%%%%%%%%%%%%%%%%%%%%%%%%%%%%%%%%%%%%%%%%%%%%%%%%%%%%%%%%%%%%
%%%%%%%%%%%%%%%%%%%%%%%%%%%%%%%%%%%%%%%%%%%%%%%%%%%%%%%%%%%%%%%%%%%%%%%%%%%%%%%

\end{document}